\documentclass[10pt,twocolumn,letterpaper]{article}

\usepackage{cvpr}
\usepackage{times}
\usepackage{epsfig}
\usepackage{graphicx}
\usepackage{amsmath}
\usepackage{amssymb}
\usepackage{blindtext}
\usepackage{caption} 
\usepackage{multirow}
\usepackage{threeparttable}
\usepackage{subcaption}
\usepackage{graphicx}
\usepackage{lipsum}
\usepackage[ruled, linesnumbered]{algorithm2e}
\usepackage{enumitem}

\usepackage[breaklinks=true,bookmarks=false]{hyperref}
\hypersetup{
    colorlinks=true,
}
\cvprfinalcopy 

\usepackage{authblk}

\makeatletter
\renewcommand\AB@affilsepx{\qquad \protect\Affilfont}
\makeatother

\begin{document}

\title{\vspace{-1.5cm}Towards Solving the DeepFake Problem : An Analysis on \\Improving DeepFake Detection using Dynamic Face Augmentation}
\author[$1\dag$]{Sowmen Das}
\author[$2||$]{Selim Seferbekov}
\author[$3\dag$]{Arup Datta}
\author[$4\mathparagraph$]{Md. Saiful Islam}
\author[$5\mathsection$]{Md. Ruhul Amin}

\affil[\space]{\quad $^\dag$Shahjalal University of Science and Technology, Bangladesh}
\affil[\space]{\quad $^{||}$Mapbox}
\affil[\space]{\hspace{4em}$^\mathparagraph$University of Alberta, Canada}
\affil[$\mathsection$]{Fordham University, USA \protect \\}
\affil[ ]{\hspace{-30pt} \textit {\normalsize $^1$sowmendipta@gmail.com \quad $^2$selim.sef@gmail.com \quad $^4$mdsaifu1@ualberta.ca \quad $^5$mamin17@fordham.edu}}

\maketitle


\begin{abstract}

   The creation of altered and manipulated faces has become more common due to the improvement of DeepFake generation methods. Simultaneously, we have seen detection models' development for differentiating between a manipulated and original face from image or video content. In this paper, we focus on identifying the limitations and shortcomings of existing deepfake detection frameworks. We identified some key problems surrounding deepfake detection through quantitative and qualitative analysis of existing methods and datasets. We found that deepfake datasets are highly oversampled, causing models to become easily overfitted. The datasets are created using a small set of real faces to generate multiple fake samples. When trained on these datasets, models tend to memorize the actors' faces and labels instead of learning fake features. To mitigate this problem, we propose a simple data augmentation method termed Face-Cutout. Our method dynamically cuts out regions of an image using the face landmark information. It helps the model selectively attend to only the relevant regions of the input. Our evaluation experiments show that Face-Cutout can successfully improve the data variation and alleviate the problem of overfitting. Our method achieves a reduction in LogLoss of 15.2\% to 35.3\% on different datasets, compared to other occlusion-based techniques. Moreover, we also propose a general-purpose data pre-processing guideline to train and evaluate existing architectures allowing us to improve the generalizability of these models for deepfake detection.
   

\vspace{-5pt}
\end{abstract}

\section{Introduction}

\begin{figure}
\centering
\includegraphics[width=0.99\linewidth]{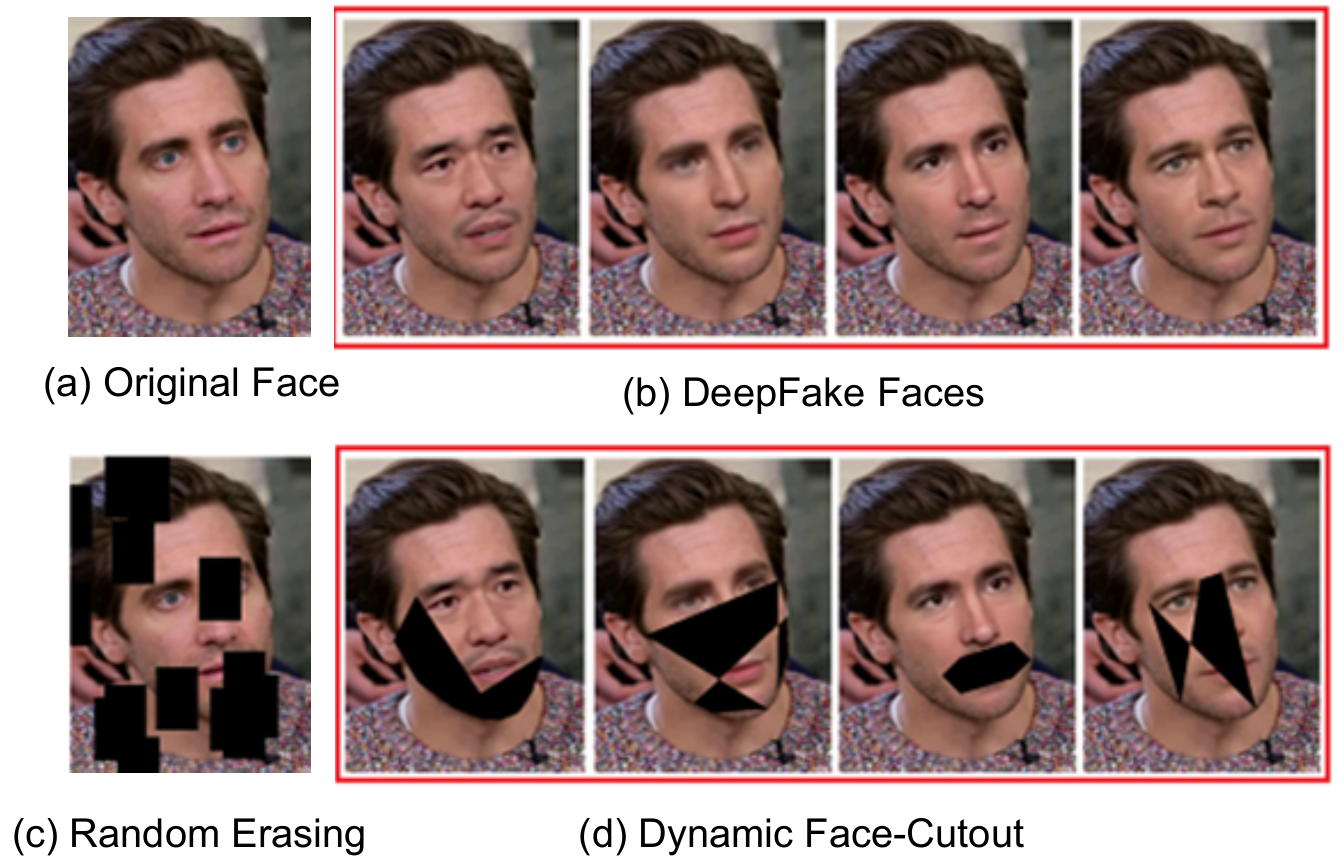}
\caption{(a) \& (b) Example of multiple DeepFake images created from an original face. (c) Random Erase augmentation. (d) Results of Dynamic Face-Cutout. Compared to Random-Erasing, our method augments a face based on the facial landmarks irrespective of its orientation and can identify manipulated facial features. \vspace{-10pt}}
\label{fig:intro}
\end{figure}

The term \textit{DeepFake} has gained much attention in recent times. It denotes manipulated multimedia content, specifically video or images created using deep learning techniques. Deepfakes are typically created using generative networks like Variational Autoencoders (VAE) \cite{vae}, and Generative Adversarial Networks (GAN) \cite{gan}.
Although digital media manipulation is not new, the use of deep learning architectures for this purpose has been gaining popularity.
Deepfake forgeries work by altering the facial attributes \cite{hao2020fargan}, gestures \cite{chan2018everybody}, or swapping the entire face of a target subject \cite{faceswap}. A deepfake generator first learns the facial features and attributes of the subjects and then generates a forged face by selectively altering these attributes. 

The application of machine intelligence systems has enabled deepfake generators to produce forgeries almost unidentifiable by human inspection. As such, deepfake detection is one of the significant challenges of digital forensics and media security. Deepfakes pose a considerable risk to the authenticity and security of the current information media. They can be used as tools of political propaganda, spreading misinformation, identity fraud, and blackmail. Deepfakes have exposed the domain of machine intelligence to ethical risks and are the prime example of the harmful impacts of current AI systems.


In response to the increasing development of deepfake generators, many actions are being taken to develop techniques for detecting these forgeries. Institutes like Facebook, Google, and DARPA have independently made efforts to this end by releasing large-scale datasets and organizing benchmark competitions. 
As a result of these efforts, a large number of deepfake detection approaches \cite{li2020face, haliassos2020lips, sun2020identifying, tarasiou2020extracting, tolosana2020deepfakes, guo2021fake}, as well as a number of datasets \cite{dfd,dfdcresult,headposes,li2020celeb,jiang2020deeperforensics,rossler2019faceforensics++} have been proposed in recent times. 

Deepfake detection is considered a binary classification problem, where given an image or video containing a face, the detector needs to identify whether the face is real or forged. Existing datasets are composed of collections of videos and images of real and fake faces. Typically the detection networks are trained on these images to learn to identify deepfake features. After careful analysis of these existing networks and the results from the benchmark challenges, we have identified that most of these detection methods do not generalize well to external perturbations \cite{dfdcresult}. Even after training on such large amounts of data, these methods generally fail when tested on external samples. The winning solution that was proposed in the Kaggle DFDC competition by \textit{Selim Seferbekov} \cite{dfdcresult} utilized an EfficientNet \cite{tan2019efficientnet} model to score a LogLoss of $0.20336$ on the public test dataset. This shows that existing vision architectures have the ability to understand and identify deepfake features. However, this score also signifies that even after training with such a huge amount of data, the model is still not robust enough to be used for real-time attributions. The organizers of DFDC have stated that \textit{``Deepfake detection is still an unsolved problem''} \cite{dfdcresult}.

\medskip
\noindent \textbf{Motivation: } The main motivation of our research is to identify the causes of performance degradation for existing networks. We have formulated the problems surrounding the current state of deepfake detection into two questions.
\vspace{0.2em}
\setlist{nolistsep}
\begin{enumerate}[leftmargin=*]
    \item Why can deepfake detection models trained on large amounts of data not generalize well to external perturbations?
    \vspace{0.2em}
    \item How can we use existing vision architectures for the purpose of deepfake detection unbiased to external dataset?
\end{enumerate}
\vspace{0.2em}

\noindent To answer the first question, we take a deeper look at the data used to train these existing architectures in Sec. \ref{sec:data}. All currently available public deepfake datasets include a collection of real and manipulated videos and images. In general, dataset creators first collect some videos of real faces and apply different deepfake generators on those faces to create many fake face samples. A problem with this process is that it is quite challenging to find a consistent amount of real faces along with the consent related to the use of such facial data. Datasets like UADFV \cite{headposes}, FaceForensics++ (FF++) \cite{rossler2019faceforensics++} and Celeb-DF \cite{li2020celeb} were created using videos collected from online video streaming sites like YouTube. On the other hand, Google DFD \cite{dfd}, and the DFDC dataset was created using video clips recorded by paid actors.
To mitigate the scarcity, in most of these datasets a \textit{single} authentic face is over-sampled to generate multiple fake samples. This lack of variation and oversampling of data causes neural networks to quickly overfit the data before learning the necessary features for deepfake identification. 
Since detection models are trained with a handful of unique faces, they start to memorize the subjects' faces and their corresponding labels resulting in poor generalization.

To answer the second question, we propose a data augmentation method termed as \textit{Face-Cutout} to train the existing vision architectures on the oversampled deepfake datasets.
Our proposed method
is a variation of the existing Random-Erasing \cite{randomerase} augmentation that repaints groups of pixels of different shapes on an image using face landmark information.
Random-Erasing is a popular and effective augmentation that employs the cutout method by selecting rectangles of various sizes and replacing the pixels with random values.
This augmentation has the added effect of Dropout \cite{dropout} regularization, which is highly effective in reducing overfitting. However, we have found in our experiments that plain Random-Erasing can be detrimental to deepfake detection as it does not consider the underlying pixel information of the erased segment. 
To improve the functionality of Random-Erasing, and utilize it for deepfake detection, Face-Cutout uses the face landmark information and also the deepfake locations to dynamically select the best cutout region, as can be seen from Fig. \ref{fig:intro}. It provides selective attention by occluding areas of the face that do not contain fake information. 
Since our method can improve the variation of the datasets, we can successfully train existing general-purpose vision architectures without worrying about them becoming overfitted to the training samples.

\medskip
\noindent \textbf{Contributions:} The contributions of our work focus on identifying the problems of bias and data variance of existing deepfake datasets and understand their shortcomings. The deepfake phenomenon is itself a significant challenge for the security of any modern AI system. It is the responsibility of efficient pattern recognition methods to distinguish the authenticity and reliability of any media content. We aim to propose an effective solution to tackle the challenges of detecting deepfakes by utilizing existing detection frameworks and improving their performance. 
To summarize our contributions,
\vspace{0.2em}
\setlist{nolistsep}
\begin{itemize}[leftmargin=*]
    \item We provide a comprehensive analysis of popular deepfake datasets to identify their shortcomings. 
    \vspace{0.2em}
    \item We show the use of face clustering for evaluating the datasets and propose its use as a general data pre-processing step to prevent data leaks.
    \vspace{0.2em}
	\item We propose \textit{Face-Cutout}, a simple erasing technique using the facial landmarks and underlying image information to dynamically cutout regions for augmentation.
	\vspace{0.2em}
    \item We show that our technique significantly improves deepfake detection performance of existing architectures by reducing overfitting and improves generalization.
\end{itemize}


\section{Related Works}
\label{sec:related_works}

Most state-of-the-art deepfake generators utilize GANs for face forgery. The GAN is trained with a dataset of face images to learn the identifying characteristics and attributes of a face. During generation, these attributes can be selectively modified to produce a different face. Multiple variations of GAN architectures \cite{stylegan, choi2017stargan, CycleGAN2017, pix2pix} are currently used for deepfake generation. They are capable of performing image-to-image translation \cite{nirkin2019fsgan}, modify the age and gender of a face \cite{liu2019stgan}, and swap faces of two persons \cite{zhang2019faceswapnet}. An extensive study of DeepFake generation is presented in \cite{gansurvey}. 

For detecting deepfake content, several different architectures have been proposed so far. Shallow networks were proposed in \cite{afchar2018mesonet} to exploit the mesoscopic features in DeepFake videos having high compression artifacts. XceptionNet has been shown to perform very well in identifying facial forgeries in \cite{rssler2019faceforensics}. A Siamese approach is used in \cite{siamese} that learns a difference function from both real and manipulated frames to encode deepfake features. In \cite{nguyen2019multitask} multi-task learning was carried out to simultaneously classify and localize deepfakes. The use of capsule networks for forgery detection has been highlighted in \cite{capsule}. In \cite{onthedetection} a visual attention network with supervised and unsupervised learning approach has been shown. In addition, many studies have been done to detect the various artifacts left by deepfake generators, such as face warping artifacts \cite{fwa}, temporal and spatial inconsistencies \cite{rnn}, eye blinking \cite{oculus}, inconsistent head poses \cite{headposes}, \textit{etc}. In recent times, attention and transformer networks have also been used in different methods \cite{zhao2021multiattentional, Kim_2021_CVPR,wodajo2021deepfake,wang2021m2tr,heo2021deepfake}.  A survey of multiple DeepFake detection architectures and their performance comparison has been presented in \cite{survey21}.

Although several studies have been done on designing architectures and identifying feature descriptors for fake face classification, there has been little analysis of the effects of data pre-processing and training specifics. The single pre-processing step used in all image-based detectors is extracting video frames and locating the facial regions. The effect of augmentations for face detection and CNN training has been studied in \cite{bondi2020training, faceprocess} In addition to face detection, \cite{rnn} adopted face tracking and face alignment. In \cite{li2020face} a novel image representation has been presented that uses facial landmarks to generate a blending mask before training. This method tries to determine the boundary of manipulated facial regions created due to the blending of different faces. 

Overfitting is one of the main challenges in training neural networks. Dropout \cite{dropout} is one of the most widely used regularization techniques that is used o mitigate this problem. These layers are inserted in-between any standard convolution layers where they randomly drop the activations of a fixed amount of neurons. Since different neurons are deactivated at each iteration, the network cannot become over-confident on a fixed feature set. However, as dropout layers work directly on the feature activations, they cannot be tuned. Random-Erasing augmentation introduces the effects of dropout outside the network. Random-Erasing selects random rectangular groups of pixels from an image and removes them to generate an occluded sample. The process is similar to dropout because it randomly deletes some specific features. However, this approach has a drawback in that randomly cutting out patches can remove essential object descriptors. For instance, if the upper half of an image containing the number \textit{`8'} is removed, it becomes a \textit{`6'}. In such cases, random erasing is detrimental to the training process. This effect is much more visible for deepfake images as the number of fake pixels is sometimes so low that random erasing can remove them entirely. So, to fix this problem and apply this regularization for deepfakes, we introduce our improved Face-Cutout augmentation.

\begin{figure}[!b]
\centering
\includegraphics[width=0.95\columnwidth]{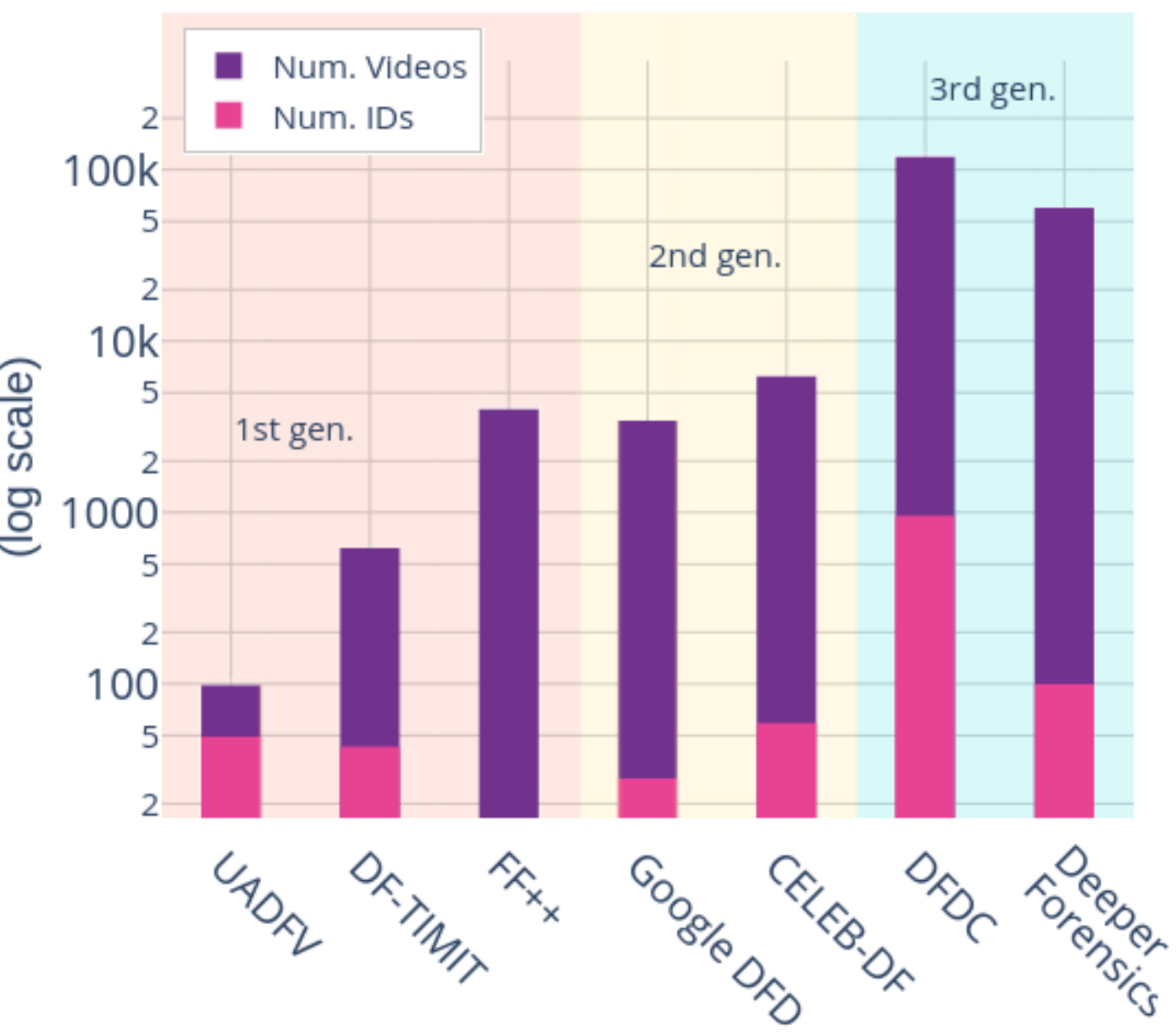}
\caption{Comparison of current DeepFake datasets. Y-axis is shown in the log scale since the DFDC dataset is over an order of magnitude larger than any others. The number of videos and  IDs is overlaid on the graph to show a comparison of the ratio of ID:Video. Datasets are divided into generations as given in \cite{li2019celebdf}.}
\label{fig:datasets}
\end{figure}

\begingroup
\setlength{\tabcolsep}{5pt} 
\renewcommand{\arraystretch}{1.2} 
\begin{table*}
\captionsetup{font=small}
\centering
\resizebox{\linewidth}{!}{%
  \begin{threeparttable}
     \begin{tabular}{c|c|c|c|c|c|c|c|c||c|c|c|c} 
\hline
\multirow{2}{*}{Dataset~} & \multirow{2}{*}{\begin{tabular}[c]{@{}c@{}}Frames\\per video \end{tabular}} & \multicolumn{2}{c|}{Real} & \multicolumn{2}{c|}{Fake} & \multirow{2}{*}{\begin{tabular}[c]{@{}c@{}}Fakes per\\ Real video \end{tabular}} & \multirow{2}{*}{\begin{tabular}[c]{@{}c@{}}Unique\\ Subjects \end{tabular}} & \multirow{2}{*}{\begin{tabular}[c]{@{}c@{}}Videos \\ per subject \end{tabular}} & \multirow{2}{*}{\begin{tabular}[c]{@{}c@{}}No. of\\Clusters\\ \end{tabular}} & \multicolumn{3}{c}{Videos per cluster}  \\ 
\cline{3-6}\cline{11-13}
                          &                                                                             & \#videos & \#frames       & \#videos & \#frames       &                                                                                  &                                                                             &                                                                                         &                                                                              & min & max & avg                         \\ 
\hline \hline
FF++{\footnotesize\textsuperscript{*}}                      & {\raise.17ex\hbox{$\scriptstyle\sim$}}500                                                                         & 1,000    & 509.9k         & 4,000    & 1.7M           & 1:4                                                                              & -                                                                           & -                                                                                       & 987                                                                          & 1   & 5   & 1                           \\
Celeb-DF                  & {\raise.17ex\hbox{$\scriptstyle\sim$}}382                                                                         & 590      & 225.4k         & 5,639    & 2.1M           & 1:10                                                                             & 59                                                                          & 1:182                                                                                   & 45                                                                           & 5   & 57  & 14                          \\
DFDC                      & 300                                                                         & 19,154   & 5.7M           & 99,992   & 29.9M          & 1:5                                                                              & 960                                                                         & 1:124                                                                                   & 866                                                                          & 25  & 345 & 33                          \\
\hline
\end{tabular}
    \begin{tablenotes}
    {\footnotesize \item \textsuperscript{*} Subject data was not published}
    \end{tablenotes}
  \end{threeparttable}
  }
  \caption{Quantitative comparison and result of face clustering on various DeepFake datasets.} 
  \label{tab:quant}
\end{table*}

\begin{figure*}[!ht]
	\begin{subfigure}[b]{0.5\textwidth}
		\centering
		\includegraphics[width=\textwidth]{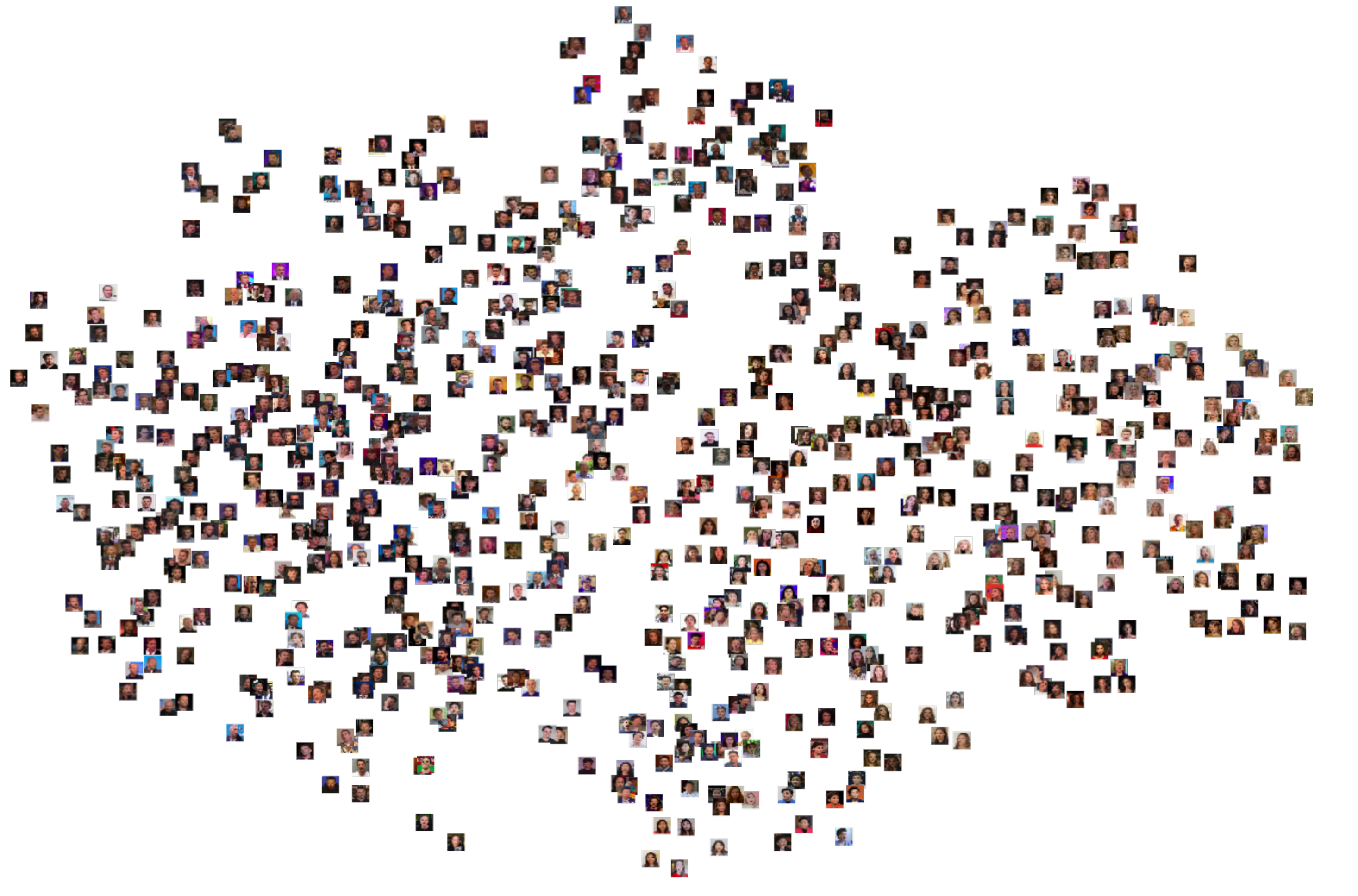}
		\caption{Celeb-DF Dataset}
	\end{subfigure}%
	\hfill
	\begin{subfigure}[b]{0.5\textwidth}
		\centering
		\includegraphics[width=\textwidth]{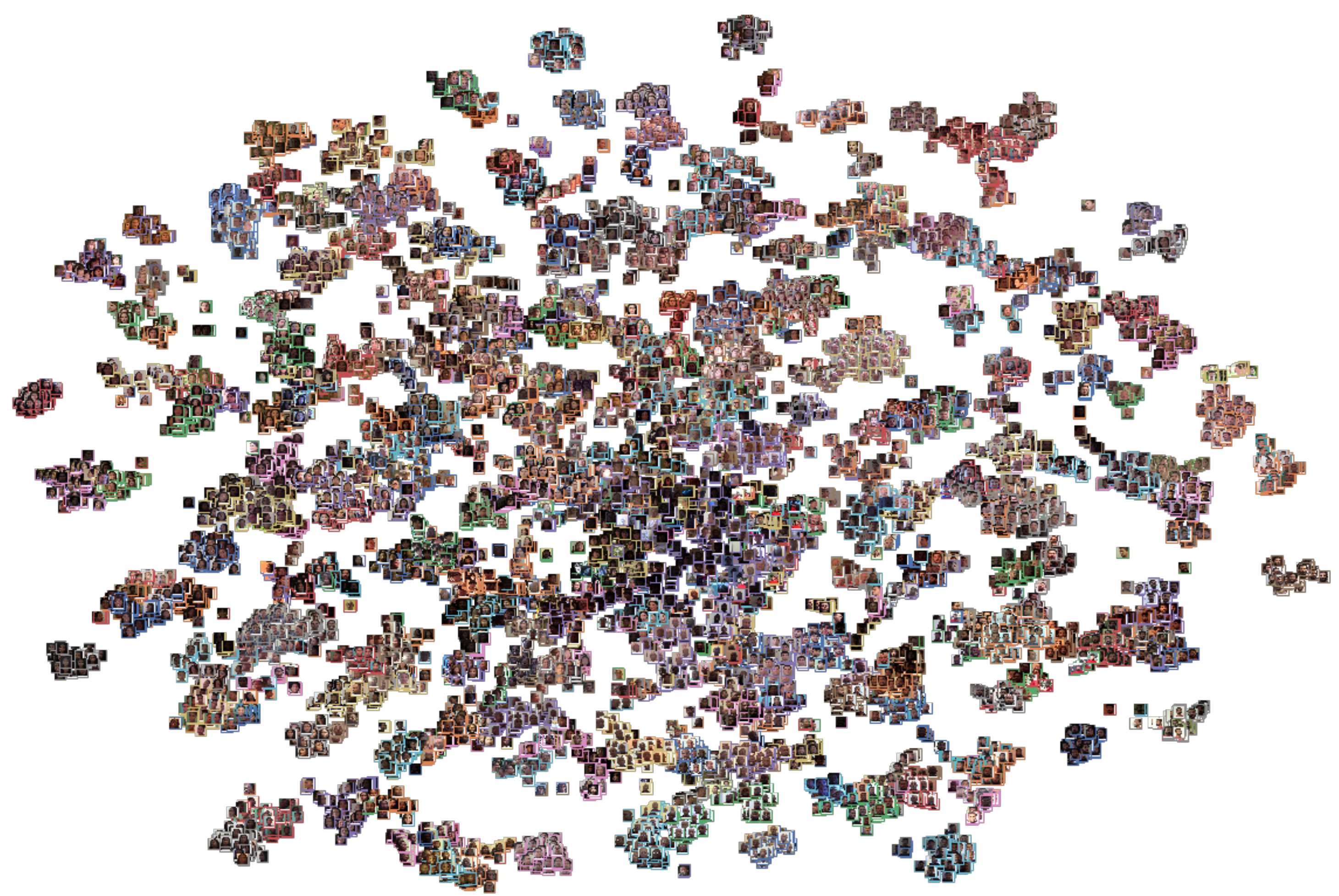}
		\caption{DFDC Dataset}
	\end{subfigure}
	\par \medskip 
	\centering
	\begin{subfigure}[b]{0.3\textwidth}
		\centering
		\includegraphics[width=\textwidth]{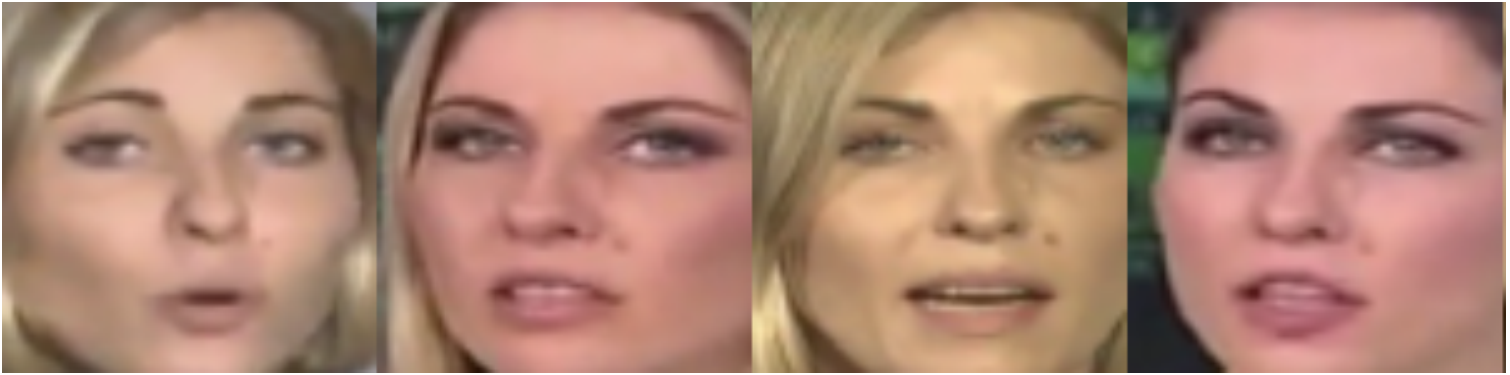}
	\end{subfigure}%
	\hspace{1em}%
	\begin{subfigure}[b]{0.3\textwidth}
		\centering
		\includegraphics[width=\textwidth]{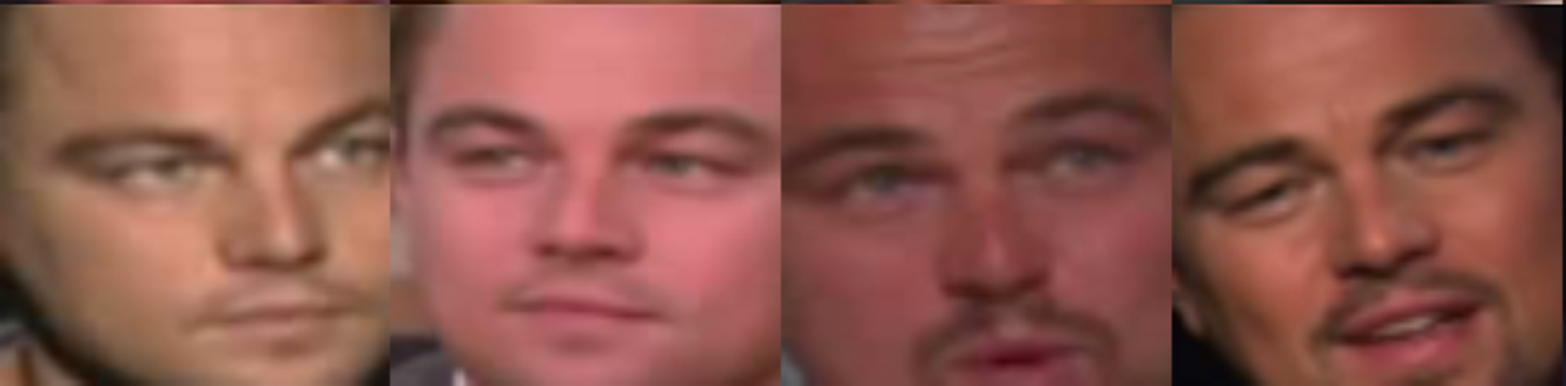}
	\end{subfigure}%
	\hspace{1em}%
	\begin{subfigure}[b]{0.3\textwidth}
		\centering
		\includegraphics[width=\textwidth]{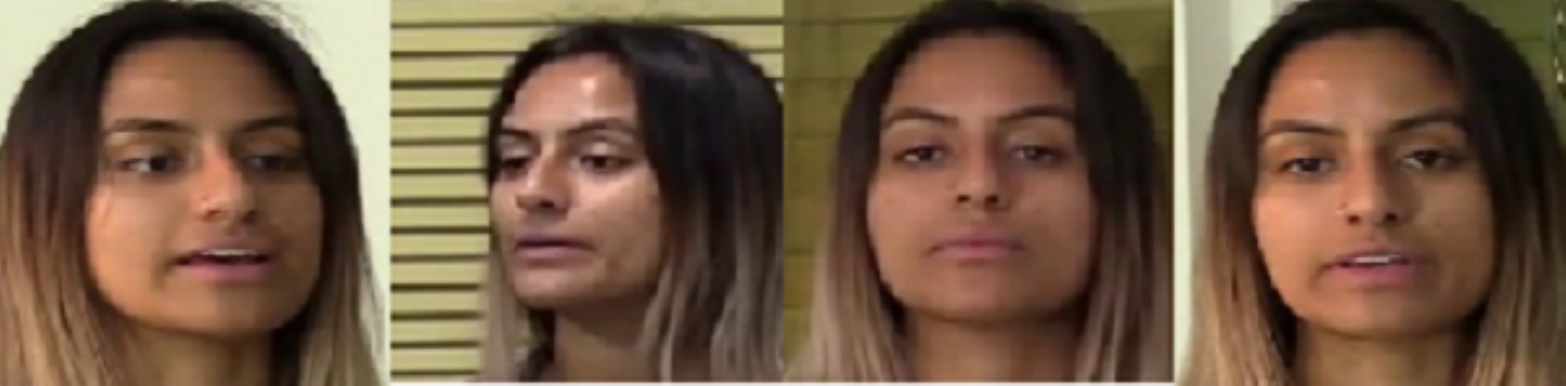}
	\end{subfigure}
	\par \vspace{0.3em}
	\begin{subfigure}[b]{0.3\textwidth}
		\centering
		\includegraphics[width=\textwidth]{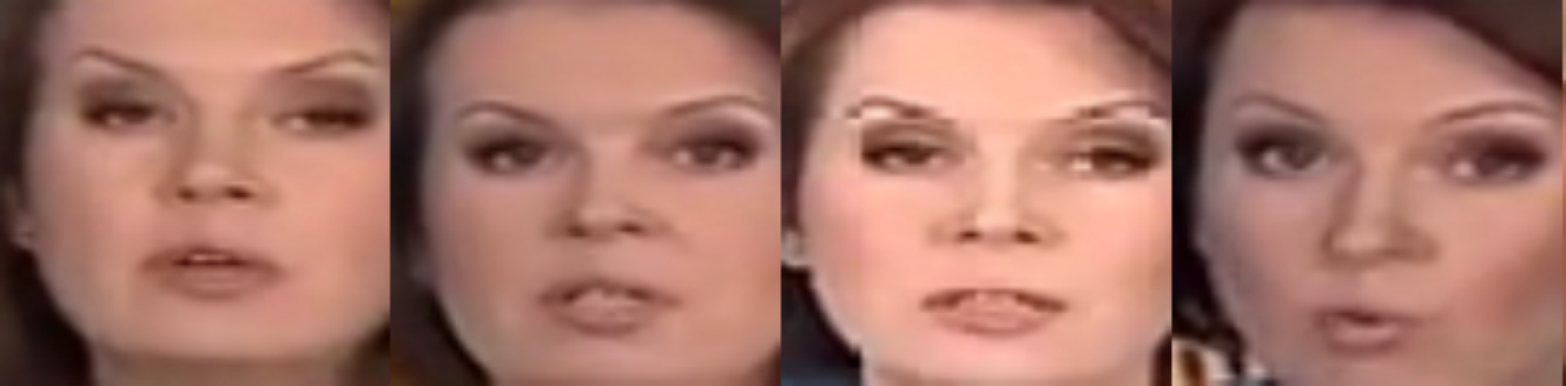}
		\caption{FF++ samples}
	\end{subfigure}%
	\hspace{1em}%
	\begin{subfigure}[b]{0.3\textwidth}
		\centering
		\includegraphics[width=\textwidth]{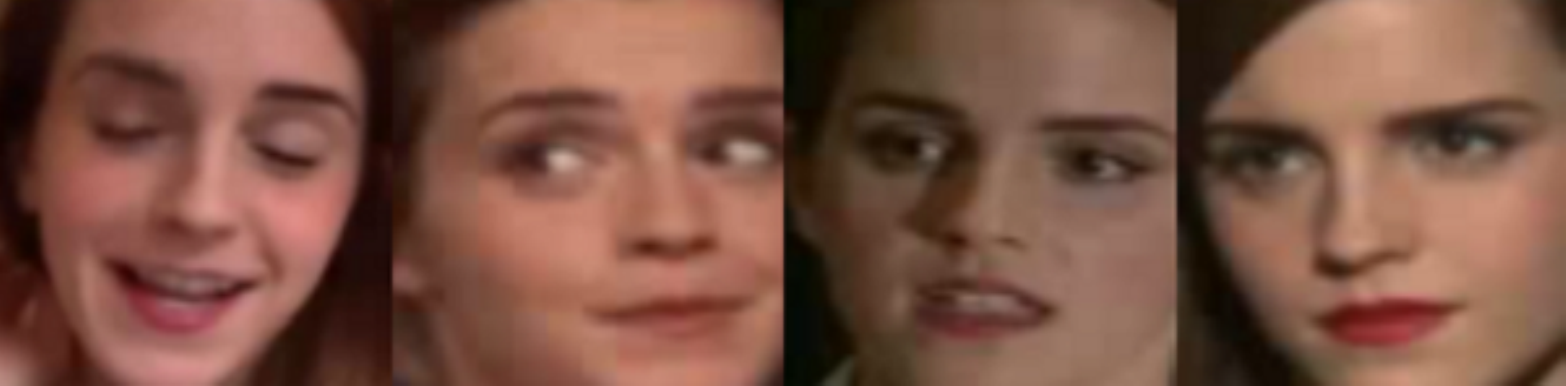}
		\caption{Celeb-DF samples}
	\end{subfigure}%
	\hspace{1em}%
	\begin{subfigure}[b]{0.3\textwidth}
		\centering
		\includegraphics[width=\textwidth]{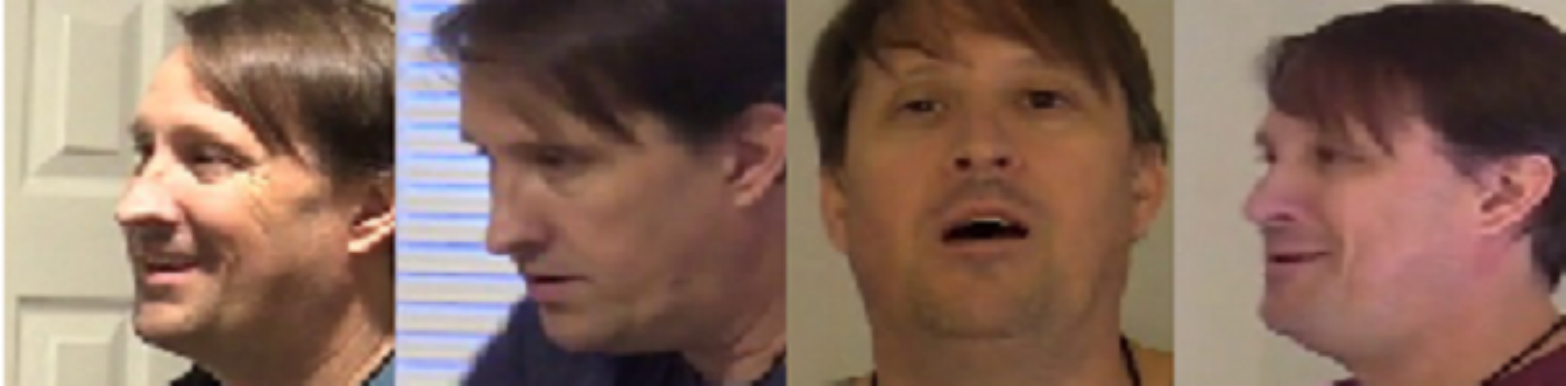}
		\caption{DFDC samples}
	\end{subfigure}
	\caption{(a) \& (b) shows the face clusters generated using DBSCAN for Celeb-DF and DFDC datasets, respectively. The density of the plots is representative of the number of videos in each dataset. (c), (d) \& (e) are sample images from clusters for each dataset. All images in these examples are from separate videos.}
	\label{fig.cluster}
\end{figure*}

\section{Identifying Dataset Issues}
\label{sec:data}

The first pressing question we needed to answer was $-$ ``\textit{Why can deepfake detection methods not generalize even after training on such large amounts of data?}" To find a solution to this question, we start by taking a closer look at the data used to train these models. A handful of deepfake datasets have been published in recent years. For creating a dataset, one needs to collect a number of unique source videos on which manipulations are applied. However, it is challenging to manually gather many unique actors and apply manipulations to each one. So generally, dataset creators select a few unique faces and generate multiple fakes from them to create a large number of videos.

We compare some of the existing deepfake datasets based on their year of release, the number of data samples, and the number of unique identities, as shown in Fig. \ref{fig:datasets}. In the figure, the number of ids represents the number of unique actor identities or unique authentic faces that were used for the dataset generation. We can see that the earlier datasets like UADFV and DF-TIMIT \cite{dftimit} had less than $500$ unique faces. As the generation improves, the amount of data is increasing exponentially. The first large scale face manipulation dataset released was the FF++ \cite{rssler2019faceforensics} benchmark dataset.
Currently, DFDC is the largest dataset in terms of both the number of videos and unique IDs. For our study, we chose to analyze the DFDC, FF++, and Celeb-DF datasets. These are the most popular datasets from their respective generations, and they represent the common trend of current deepfake data production. 

Table \ref{tab:quant} shows a quantitative comparison of the selected three datasets. The fake to real video ratio was calculated by averaging the fake video count generated from a single real source video. Average videos per subject were calculated for every subject used either as a source or target for face swap or DeepFake. We can see a large imbalance in the data distribution from the count of videos per subject. For both the DFDC and Celeb-DF datasets, a single face appears in more than $120$ videos on average. Considering the total number of frames in DFDC, a single face can be found in {\raise.17ex\hbox{$\scriptstyle\sim$}}37.2k images. So, this clearly indicates that existing datasets are highly oversampled.

\subsection{Face Clustering}
\label{sec.cluster}

To reiterate our findings regarding face oversampling, we further perform a clustering of the unique faces in the datasets to visualize the extent of the problem. The first step in face manipulation detection is to locate the face from a video frame. Several different face detection models are used for this purpose, including, MTCNN \cite{mtcnn}, DLib \cite{dlib}, BlazeFace \cite{bazarevsky2019blazeface} etc. We used the MTCNN face detector to detect, and cluster faces from videos. 
We use facial clustering to aggregate similar faces and calculate the number of unique videos per actor.
First we extracted all faces found in each \textit{real} video and encoded each face to a $128$ dimension vector using a one-shot CNN encoder \cite{facerec}. Experiments showed that increasing the dimension size did not improve cluster estimation by any significant margin. Then we used Density-Based Spatial Clustering of Applications with Noise (DBSCAN) \cite{dbscan} with Euclidean distance to congregate the images into groups. Using the labeled dataset, we assigned all fake videos to their respective source face cluster. The clusters for Celeb-DF and DFDC are shown in Fig. \ref{fig.cluster}. The $128$ dimensional embeddings have been reduced to $2$ dimensions using Principal Component Analysis (PCA) for better visualization. Here, each cluster corresponds to a unique face. We can see that even though there are many videos, unique clusters are very low. 

Table \ref{tab:quant} contains the result of clustering. We can see that the algorithm identified $45$ clusters in the Celeb-DF dataset among the $59$ subjects reported by the authors. The difference is due to the difficulties of differentiating among actors of similar race and gender under different lighting conditions and low-resolution images extracted from compressed videos. For FF++, we identified $987$ clusters. This means almost all real videos in this dataset have unique faces. For the DFDC dataset, we identified $866$ separate clusters compared to $960$ reported subjects. The variety of faces in DFDC is meager compared to its size. 

The results of face clustering further solidify our claim concerning the data variation. So, from this data and our observation, we can conclude that the reason for models not being able to generalize well to external cases is because the available training data is heavily oversampled. Models trained on these datasets are easily overfitted. Moreover, this phenomenon is complicated to identify because of data leakage. So, if we can find a way to mitigate overfitting, we can train existing vision architectures on these large datasets and utilize them for deepfake detection.

\subsection{Pre-processing Guidelines}
\label{guideline}
Data leak is a phenomenon when a training sample is also used as part of the validation set. Measuring model performance on leaked validation data results in skewed and overly accurate metrics. Usually, before the start of a training routine, the available data is divided into train, validation, and test splits using random or stratified splitting techniques \cite{reitermanova2010data}. The validation and test data need to be isolated so that they are not used for training. Because if a model is trained and evaluated on the same data, we will not be able to identify overfitting or the robustness of the model. However, a significant reason why overfitting is challenging to identify in deepfake datasets is that existing splitting techniques do not work for these datasets. 

Even though the datasets contain real and fake faces, we cannot randomly split the data based on labels only. Since we have identified that the models can overfit the faces, we need to split the data based on the uniqueness of the faces. If a face used for training is also available in the test set, the model can memorize the face and predict the label resulting in high test accuracies. So we will need to split the data so that there are no common faces between the train and test splits. For this purpose, we propose the use of face clustering. The pre-processing steps are:
\vspace{0.2em}
\setlist{nolistsep}
\begin{enumerate}[leftmargin=*]
    \item Group the data based on the available unique faces. All videos or images containing the same face should be considered as a single unit.
    \vspace{0.2em}
    \item Split the data based on the number of face clusters, rather than available labels.
\end{enumerate}
We propose these pre-processing steps as a general guideline for training deepfake detection networks. This will prevent data leaks and allow researchers to identify the robustness and generalizability of their models.


\section{Face-Cutout}
\label{sec:aug}

\begin{figure}[b]
\centering
	\begin{subfigure}[b]{0.35\linewidth}
		\centering
		\includegraphics[width=\linewidth]{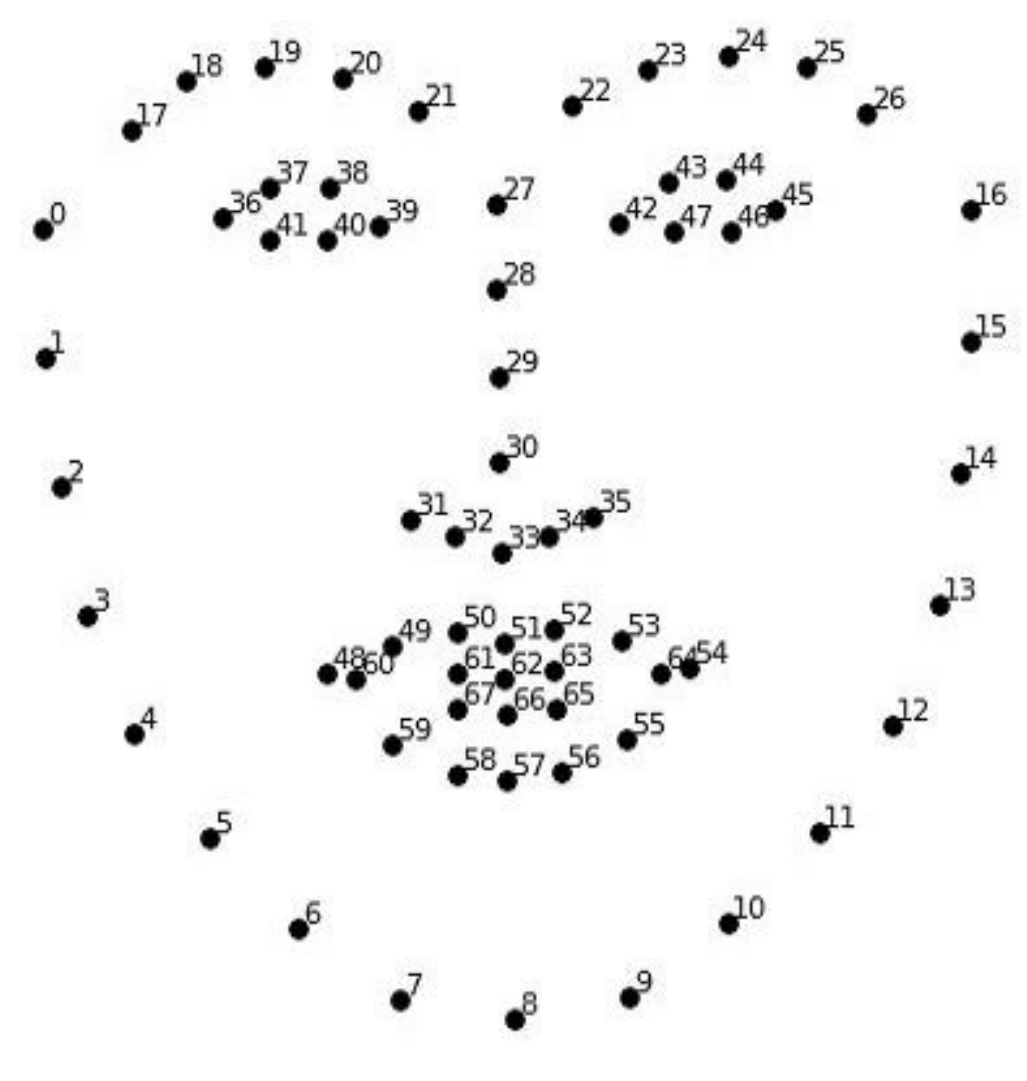}
	\end{subfigure}%
	\hspace{0.3em}
	\begin{subfigure}[b]{0.35\linewidth}
		\centering
		\includegraphics[width=\linewidth]{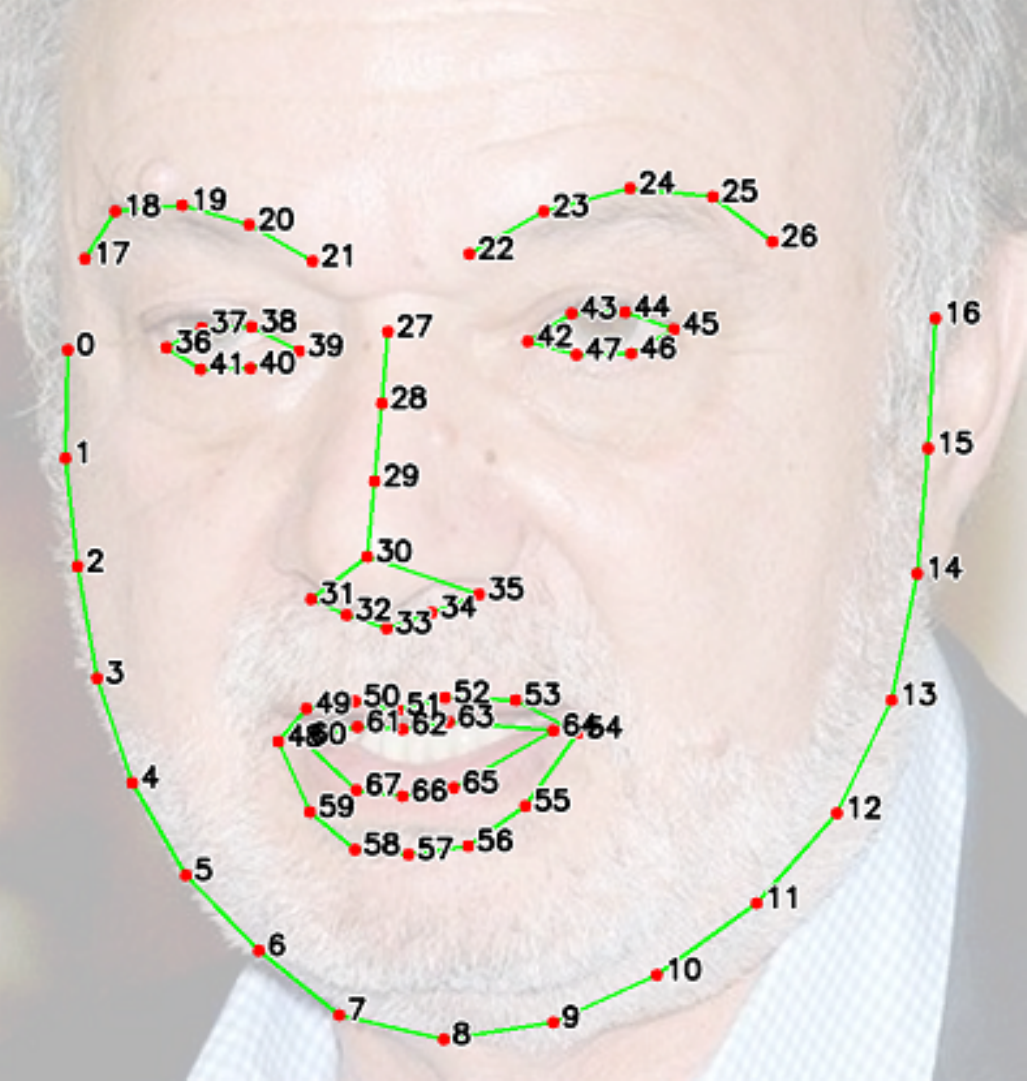}
	\end{subfigure}
	\caption{68 landmark positions detected by DLib.}
	\label{fig.landmark}
\end{figure}
\begin{figure*}
\centering
\includegraphics[width=0.98\linewidth]{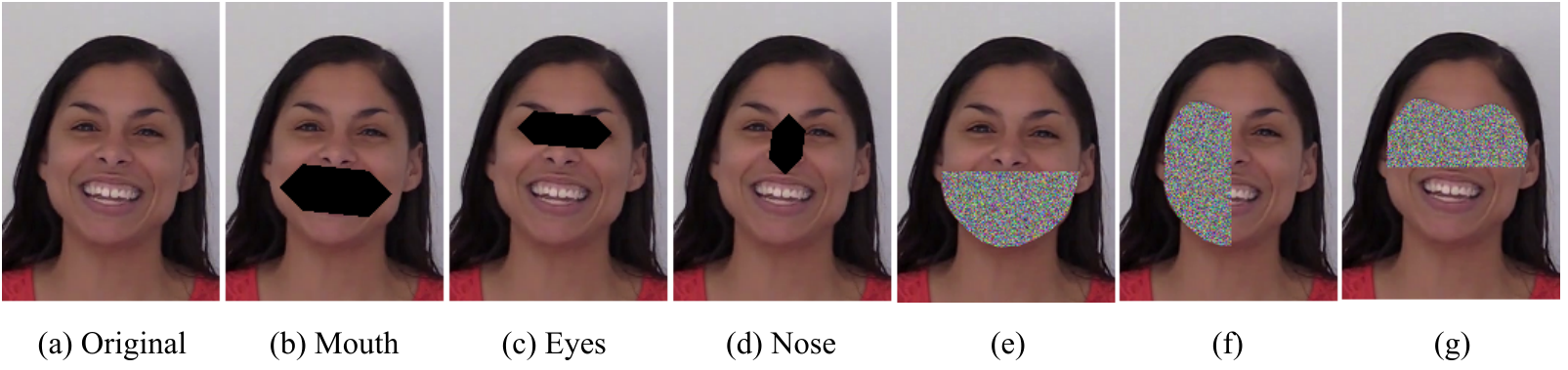}
\caption{(a) The original face without any augmentations. (b), (c), (d) Three types of Sensory Group cutout of Mouth, Eyes and Nose. (e), (f), (g) Three random outputs for Convex-Hull Cutout. It also shows example of cutout fill with random values.}
\label{fig:cutout}
\end{figure*}

\begin{figure}
\centering
\includegraphics[]{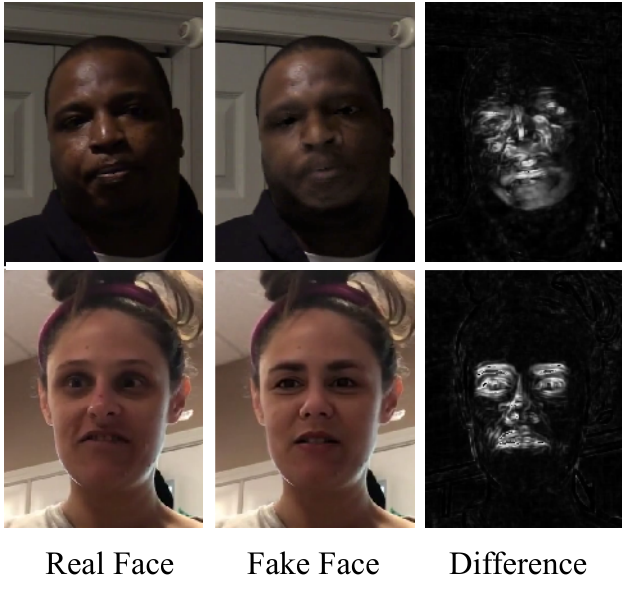}
\caption{Face extracted from the frame of a real as well as its corresponding fake video. The difference mask shows the artifacts found by measuring the real and fake face's pixel-wise difference. \vspace{-20pt}}
\label{fig:mask}
\end{figure}
Our proposed method, Face-Cutout uses the landmark positions of a face to augment training images. Landmark positions are the locations of eyes, ears, nose, mouth, and jawline. DLib \cite{dlib} can uniquely identify $68$ positions on the face, shown in Fig. \ref{fig.landmark}. Each landmark position is annotated with a point ranging from $[0-67]$. Points in $[36-41]$ and $[42-47]$ denote left and right eyes respectively, while $[27-35]$ is used for the nose and $[48-67]$ for the mouth.
We use these positions to calculate polygons for Face-Cutout. Before training, we generate a pixel-wise difference mask by calculating the Structural Similarity Index (SSIM) \cite{ssim} between the frame of a real, and it's corresponding fake video, as shown in Fig. \ref{fig:mask}. This difference mask contains $1$ for manipulated pixels and $0$ for real ones. The algorithm takes as input a face image and its corresponding mask to generate an augmented image.


For a training image, $I$ in a mini-batch, the probability of it undergoing augmentation is $p$. 
Our augmentation method has three steps: $-$ 1) Polygon proposal, 2) Polygon Selection, and 3) Polygon Filling. In the first step, we select a random group of points from the $68$ landmark coordinates and propose several polygons using these coordinates as vertices. We perform two types of cutout operations depending on the point group; 1) Sensory group removal and 2) Convex-hull removal. Sensory groups consist of polygons created using the landmark groups of two eyes, nose and mouth. Convex-hull removal uses all $68$ coordinates to generate random polygons using the convex-hull algorithm. Next, we select the maximum enclosing polygonal region $I_c$ from these proposed polygons using the pre-calculated difference mask. We are choosing the maximum region because we want to remove as much irrelevant region as possible. So, the model can focus on only the fake regions. Let $C_o$ be the set of all pixels of value $1$ contained within the selected polygon region and $A$ the set of all $1$ pixels in the entire mask. Therefore, by definition $C_o \subset A$. The amount of envelop of a proposed cutout region is denoted by $\rho$ where,
\begin{equation}
\label{eq.rho}
    \rho = \frac{\mid C_o \mid}{\mid A \mid}
\end{equation}


\noindent The polygon is selected as a cutout region if $\rho \leq \Gamma_h$ where $\Gamma_h$ is a predefined threshold set to a default of $0.3$. Since for real images, the difference mask does not contain any $1$; therefore, $|A|$ will always equal $0$. So, they are augmented by the default polygon generated by the algorithm. Thus, the condition of $\rho$ is only applied to fake images. Finally the selected polygon is cut out or filled. By cutout, we mean that the pixel values of the selected region $I_c$ are replaced with values from $[0,255]$. Face-Cutout can also be combined with any existing image augmentation, like rotation, scaling, and color transforms. Fig. \ref{fig:cutout} shows some images generated using Face-Cutout.



\section{Experimental Setup}
\label{sec:exp_setup}

We evaluate Face-Cutout on the three datasets, FF++, Celeb-DF, and DFDC. FF++ videos were used at 40\% compression, and the other two datasets were used in their original format. We also tested combining samples from all three datasets to present that our method can help models to generalize to different datasets.

\medskip
\noindent \textbf{Test Set Selection: } As explained earlier, DeepFake datasets are prone to overfitting due to a lack of face variation. While training and evaluating on these datasets, we need to prevent data leakage. For any machine learning system, the method for selecting the optimal validation data is a highly studied problem \cite{validation}. Train data should not be used for validation, as it would result in unrealistic model evaluation. To ensure data separation in our experiments, we followed the face cluster guideline proposed in Sec. \ref{guideline}. We evaluated the models with K-Fold Cross Validation with $K$ = $10$ and used a single holdout set for the test. For the combined evaluation, we selected a subset of videos from each dataset based on clusters and used them together for the train and a separate set for the test. 


\medskip
\noindent \textbf{Model Selection:} We selected two deep convolutional models; EfficientNet-B4 and XceptionNet. Both were initialized with pre-trained ImageNet weights. Additionally, EfficientNet-B4 was pre-trained using Noisy Student \cite{noisystudent}. There are eight variants of the EfficientNet architecture based on their depth and number of parameters ranging from B0-B7. We chose B4 because of it's lower parameter count and faster train time. The second model, XceptionNet, was introduced in \cite{rssler2019faceforensics, dolhansky2019DeepFake} as a baseline that achieves great results in forgery detection tasks.

\medskip
\noindent \textbf{Preprocessing:} Each dataset consists of real and manipulated videos in \textit{mp4} format. We used only facial regions from each frame for both training and inference. We picked every  10\textsuperscript{th} frame from each video. First, we used MTCNN Face detector with thresholds $(0.85, 0.95, 0.95)$ to get the face bounding boxes from each frame of the original videos. Face detection was only applied to the real faces to avoid any false positives. Since fake videos contain the same face in the same location for each frame, the face coordinates can be used to extract the fake face. After that, DLib was used on the extracted real faces to detect facial landmarks.

\begin{figure*}
	\begin{subfigure}[b]{0.33\textwidth}
		\centering
		\includegraphics[width=\textwidth]{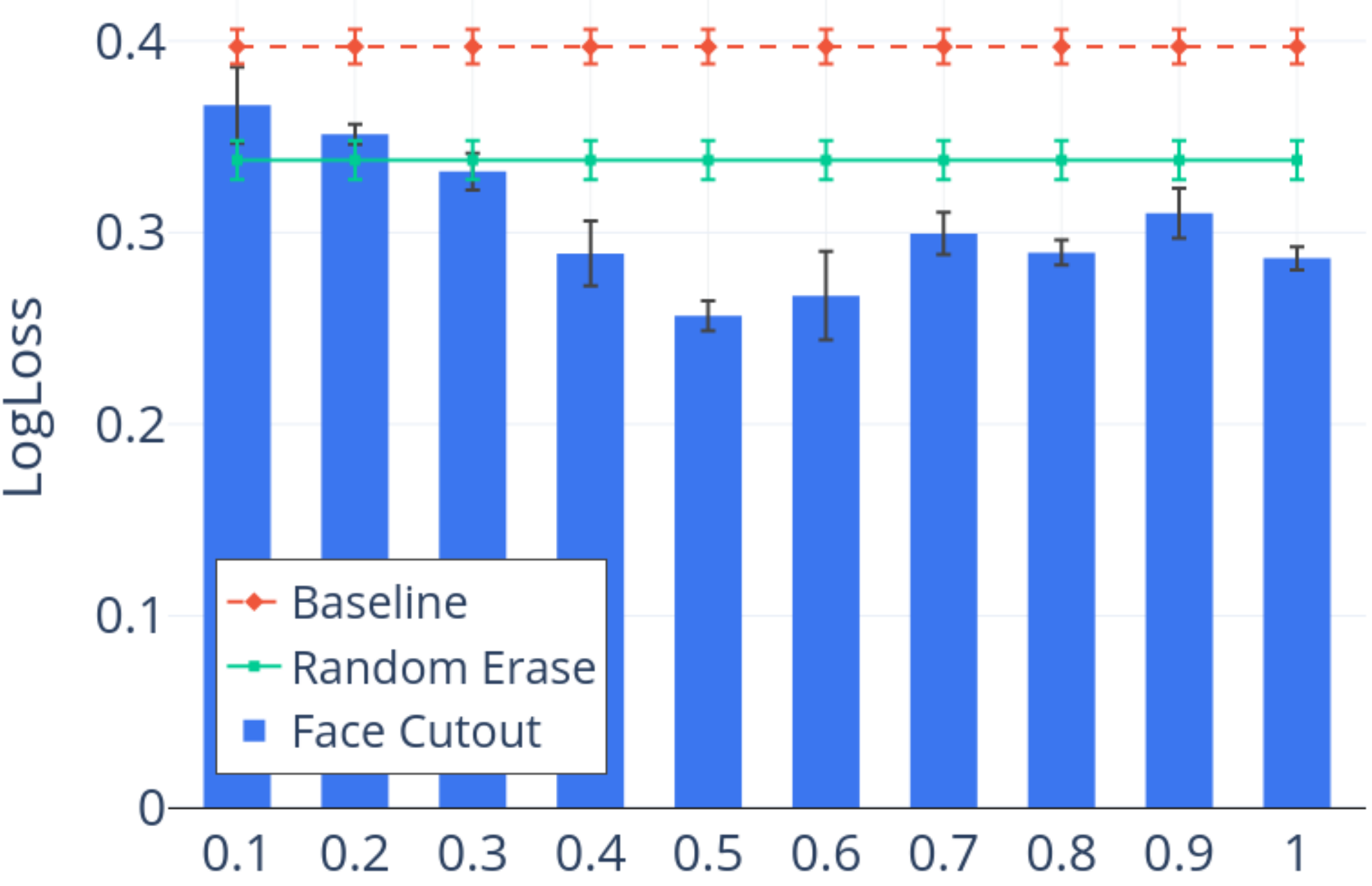}
		\caption{Probability \textit{p}}
	\end{subfigure}%
	\begin{subfigure}[b]{0.33\textwidth}
		\centering
		\includegraphics[width=\textwidth]{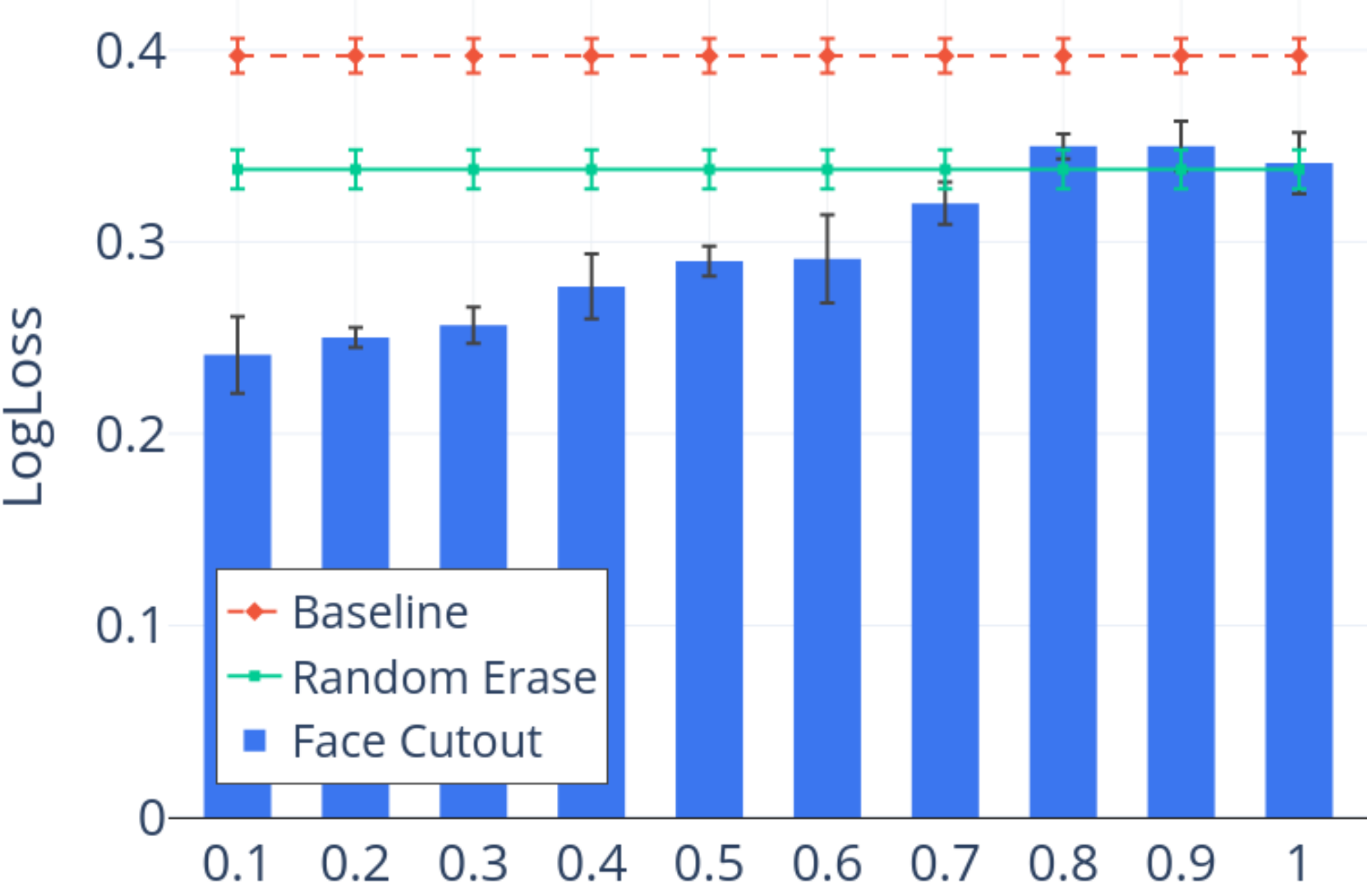}
		\caption{Threshold for $\rho$ $(\Gamma_h)$}
	\end{subfigure}%
	\begin{subfigure}[b]{0.33\textwidth}
		\centering
		\includegraphics[width=\textwidth]{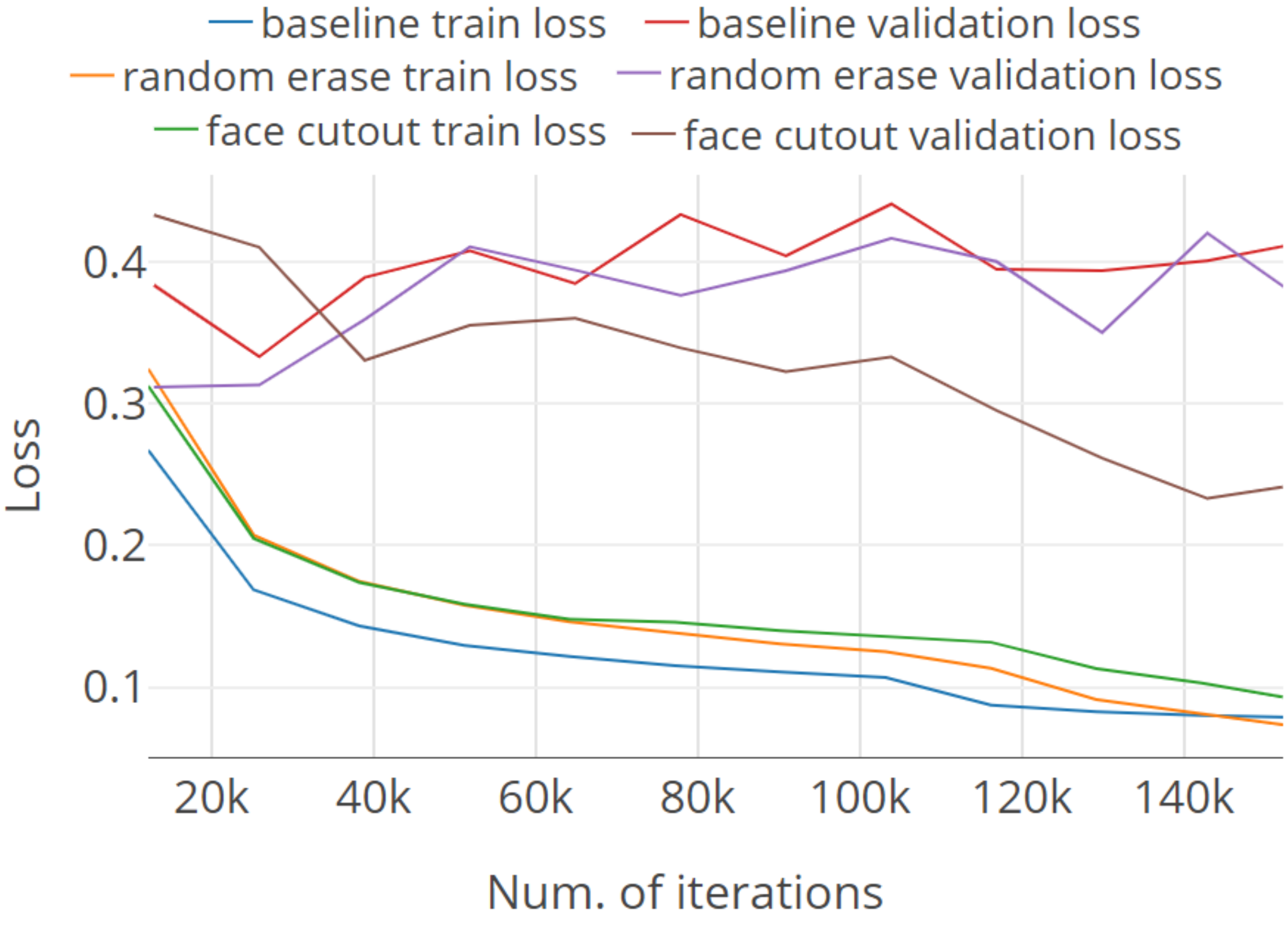}
		\caption{Validation and train losses}
		\label{fig:loss}
	\end{subfigure}
	\caption{(a), (b) Test errors under different hyper-parameters. (c) Validation and train losses on optimized hyper-parameters.}
	\label{fig.hpar}
\end{figure*}

\begin{table*}
\captionsetup{font=small}
\centering
\resizebox{0.99\linewidth}{!}{%
\begin{tabular}{l|ccc|ccc|ccc} 
\hline
\multirow{2}{*}{Model}         & \multicolumn{3}{c|}{DFDC}                            & \multicolumn{3}{c|}{FF++ (c40)}                     & \multicolumn{3}{c}{Celeb-DF}                         \\ 
\cline{2-10}
                               & LogLoss          & AUC(\%)          & mAP(\%)          & LogLoss         & AUC(\%)          & mAP(\%)          & LogLoss         & AUC(\%)          & mAP(\%)           \\ 
\hline\hline
EfficientNet-B4 Baseline       & 0.397            & 87.11           & 96.02           & 0.215           & 95.59           & 97.9            & 0.104           & 98.75           & 98.66            \\
EfficientNet-B4 + Random Erase & 0.3178           & 91.01           & 97.14           & 0.239           & 95.01           & 93.15           & \textbf{0.048}  & \textbf{99.54}  & \textbf{99.69}   \\
EfficientNet-B4 + Face-Cutout  & \textbf{0.2566}  & \textbf{92.71}  & \textbf{98.59}  & \textbf{0.178}  & \textbf{98.77}  & \textbf{99.03}  & 0.065           & 99.21           & 99.53            \\ 
\hline\hline
Xception Baseline              & 0.5598           & 78.61           & 88.51           & 0.247           & 89.91           & 95.21           & 0.199           & 98.17           & 98.05            \\
Xception + Random Erase        & 0.5011           & \textbf{82.07}  & 90.80           & 0.287           & 88.42           & 95.04           & 0.098           & 99.20           & 99.17            \\
Xception + Face-Cutout         & \textbf{0.4718}  & 81.99           & \textbf{91.32}  & \textbf{0.195}  & \textbf{96.73}  & \textbf{96.06}  & \textbf{0.096}  & \textbf{99.39}  & \textbf{99.44}   \\
\hline
\end{tabular}
}
\caption{Test results of the deep learning models trained separately using DFDC, FaceForensics++ and Celeb-DF datasets.}
\label{tab:res1}
\end{table*}

\medskip
\noindent \textbf{Training Setup:} All the extracted images in each dataset were normalized using per-channel mean of $(0.485, 0.456, 0.406)$, and standard deviation $(0.229, 0.224, 0.225)$. Images were isotropically resized to $224\times224$ with zero padding. Additional train augmentations were used, including Image Compression, Gaussian Noise, and Flipping, each with $10\%$-$15\%$ probability. Test and validation were done without any augmentation. The random generator was initialized with a seed of $777$. We used Rectified Adam \cite{radam} optimizer with an initial learning rate of $0.001$ and a weight decay of $0.0005$. Learning rate scheduling was done using Reduction on Plateau by a factor of $0.25$ and patience $2$. All models were trained using Binary Cross-entropy Loss for $30$ epochs and with Early stopping if no improvement was observed for consecutive $10$ epochs. All experimentation was conducted with a training batch size of $40$ on a system with an NVIDIA GTX 1080 Ti GPU and AMD Threadripper processor.

\medskip
\noindent \textbf{Evaluation Metrics:} Since DeepFake datasets are heavily class imbalanced, accuracy is inefficient for measuring the model performance. We decided to use Area Under Curve (AUC) of ROC and Mean Average Precision (mAP) score for analyzing model performance. The AUC score summarizes the relation between the False Positive Rate (FPR) and True Positive Rate (TPR) of our binary classifier. Moreover, since the datasets contain a larger number of true negatives, mAP is more indicative of how a detection model will perform over a real distribution of images \cite{dfdcresult}. We also measure LogLoss on videos proposed by \cite{dfdcresult} as a metric for ranking DeepFake models. It was also used for ranking submissions in the DFDC Competition. LogLoss was calculated using equation (\ref{eq:logloss}),

\begin{equation}
LogLoss =-\frac{1}{n} \sum_{i=1}^{n}\left[y_{i} \log \left(\hat{y}_{i}\right)+\left(1-y_{i}\right) \log \left(1-\hat{y}_{i}\right)\right]
\label{eq:logloss}
\end{equation}
where $n$ is the number of videos being predicted, $\hat{y}_{i}$ the predicted probability of the video being fake and $y_{i}$ the true label of the video, 1 if fake, and 0 if real. The final prediction for each video was made as a mean of each selected frame's predicted probabilities.

\section{Experimental Results}
\label{sec:result}

\medskip
\noindent \textbf{Classification on Independent Datasets:} The results of evaluating Face-Cutout on the individual datasets are shown in Table \ref{tab:res1}. We set $\Gamma_h$ = $0.3$ and cutout fill $0$. Results indicate that models trained with Face-Cutout have significant improvement over baseline and Random-Erase. Moreover, Random-Erasing performs worse than baseline in some cases, as seen from the results of FF++. Face-Cutout shows improvement in both EfficientNet and Xception models with an increase of $0.46\%$ to $7.58\%$ AUC(\%) from baseline and improvement of $15.2\%$ to $35.3\%$ test LogLoss across models and datasets. Moreover, it improved LogLoss by $19.25\%$ from Random-Erasing in DFDC and was almost on par in Celeb-DF. From Fig. \ref{fig:loss} we can see that both the baseline model and Random-Erasing overfit to the DFDC dataset considerably. The decreasing validation loss for Face-Cutout shows its effectiveness in reducing model overfitting.

\medskip
\noindent \textbf{Classification on Combined Dataset:} The combined dataset was kept small to balance the source videos since DFDC is magnitudes larger than the other two datasets. Results from Table \ref{tab:comb} show that Face-Cutout performs equally well in the combined test data and outperforms Random-Erasing. It achieves an $11.3\%$ improvement in LogLoss from Random-Erasing using EfficientNet.

\begin{table}[!ht]
\captionsetup{font=small}
\centering
\resizebox{0.99\columnwidth}{!}{%
\begin{tabular}{l|ccc} 
\hline
Model                          & LogLoss         & AUC(\%)          & mAP(\%)           \\ 
\hline
\hline
EffNet-B4 Baseline       & 0.2719          & 92.99          & 96.22           \\
EffNet-B4 + Random Erase & 0.2698          & 95.00          & 98.71           \\
EffNet-B4 + Face-Cutout  & \textbf{0.2393} & \textbf{95.44} & \textbf{98.94}  \\
 
\hline
\hline
Xception Baseline              & 0.3177          & 90.15          & 98.01           \\
Xception + Random Erase        & 0.2713          & 95.02          & 98.59           \\
Xception + Face-Cutout         & \textbf{0.2586} & \textbf{95.66} & \textbf{98.76}  \\

\hline
\end{tabular}
}
\caption{Test results on the combined dataset.}
\label{tab:comb}
\end{table}

\medskip
\noindent \textbf{Impact of Hyper-Parameters:} There are two hyper-parameters for Face-Cutout, the threshold for $\rho$ ($\Gamma_h$) and the cutout probability $p$. We experimented on the DFDC dataset using EfficientNet-B4 to measure the impact of these hyper-parameters. When evaluating one parameter, the other one was fixed. From Fig. \ref{fig.hpar}, we can see that a $p$ of around  $0.5$ achieves the lowest LogLoss and improves baseline results by a factor of $0.14$. With lower $p$, the results are close to the baseline as the augmentation isn't as effective when applied to a small amount of data. The results deviate from the optimum at higher probabilities but are still better than both baseline and Random-Erasing. For the threshold $\Gamma_h$, we can see an increasing trend in LogLoss. The threshold decides how much fake artifact we allow inside the cutout region. With higher threshold values, essential visual and fake information gets removed from the images, and the results are almost similar to Random-Erasing. A threshold of $0.1$ achieves the best score, but it has a higher error deviation. We chose $0.3$ to allow more augmentations. For Random-Erasing, we used parameters as suggested in \cite{randomerase}.

\medskip
\noindent \textbf{Impact of Cutout Pixel Value:} We evaluate Face-Cutout by erasing pixels in the selected region using three types of values: 1) each pixel is assigned a random value between $[0, 255]$, (termed as F-R); 2) all pixels are assigned with $0$, (termed as F-0); 3) all pixels are assigned with $255$, (termed as F-255). Table \ref{tab:fill} shows that all erasing schemes outperform the baseline. Moreover, F-R and F-0 performs equally well, and both are superior to F-255.

\begin{table}
\captionsetup{font=small}
    \begin{subtable}[h]{0.4\columnwidth}
        \centering
        \resizebox{0.85\linewidth}{!}{%
        \begin{tabular}[t]{l|c}
        \hline
        Fill Type   & LogLoss  \\ 
        \hline
        \hline
        F-R & \textbf{0.2547}  \\
        F-0      & 0.2566   \\
        F-255    & 0.3108   \\
        \hline
        \end{tabular}
        }
       \caption{}
       \label{tab:fill}
    \end{subtable}
    \hspace{-1.7em}
    \begin{subtable}[h]{0.7\columnwidth}
        \centering
        \resizebox{0.87\linewidth}{!}{%
        \begin{tabular}[t]{l|c} 
        \hline
        Cutout Type            & LogLoss           \\ 
        \hline\hline
        Baseline (No Cutout)               & 0.3970            \\ 
        \hline\hline
        Mouth                  & 0.3115            \\
        Eyes                   & 0.3203            \\
        Nose                   & 0.3101            \\
        Randomized Sensory (R-S) & 0.2801            \\ 
        \hline\hline
        Convex-Hull            & 0.2985            \\
        R-S + Convex-Hull  & \textbf{0.2566}   \\
        \hline
        \end{tabular}
        }
        \caption{}
        \label{tab:type}
     \end{subtable}
     \caption{(a) Test results of different fill values. (b) Results of different cutout types evaluated individually.}
     \label{tab:double}
\end{table}

\medskip
\noindent \textbf{Performance of Cutout Types:} We perform two types of Face-Cutout: 1) Sensory group and 2) Convex-Hull. We evaluated the impact of these two groups on the model performance. Table \ref{tab:type} shows that all cutout methods outperform the baseline. The use of any single sensory group cutout on all data gives approximately similar results, but choosing it randomly from the three groups improves the logloss by $0.03$. Here one notable observation is that when only the eyes were removed the logloss increased to $0.32$. This might be because eyes are vital in identification a person and during deepfake the ocular region is tampered with the most. So, removing the region entirely reduces important information for deepfake identification. Furthermore we see that using convex-hull cutout on its own performs on par with the sensory group. The best result is achieved using randomized sensory and convex-hull together.

\section{Model Interpretability }

From the different evaluation experiments, we have shown that models trained using Face-Cutout augmentation generally outperform baseline models or other augmentation methods. However, this does not entirely confirm that our proposed method is reducing overfitting. We are still not sure whether the models can identify the DeepFake features or not. To understand what the models are identifying when trained with or without the Face-Cutout, we analyze the intermediate feature representations using GradCAM \cite{cam}.

We visualize the CAM output of the EfficientNet model in Fig. \ref{fig.cam}. The baseline model identified both the sample images correctly as fake with a $98\%$ accuracy. However, the CAM output of the baseline model in Fig. \ref{fig.cam} (d) shows the model highlighting the entire face, including arbitrary parts of the image. In the activation heatmaps, brighter colors represent higher activation. The baseline model is activating around arbitrary regions of the image. So, we can state that the baseline model has been overfitted. Instead of identifying the correct fake regions, it has memorized the entire input image as fake. However, the output produced by the same model when trained with Face-Cutout as shown in Fig. \ref{fig.cam} (e) using the same parameters highlights only the fake portions of the face. This is verified by comparing the activation maps with the difference mask. So, this time the model was successful in locating the correct fake regions in the image. This means that the model trained with Face-Cutout has not overfitted to the data. So we can conclude that we have successfully reduced overfitting and improved model representation.

\begin{figure}[!h]
\centering
\includegraphics[width=0.98\linewidth]{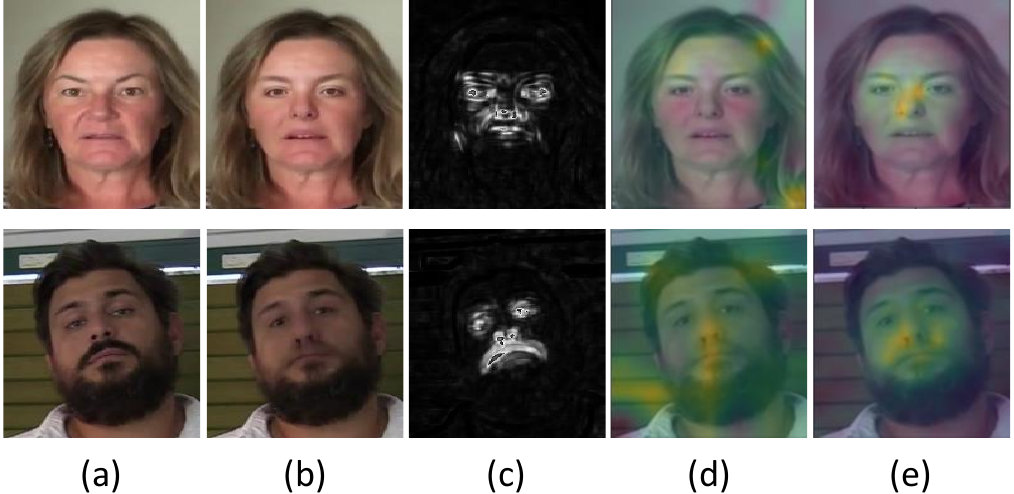}
\caption{(a) Real face, (b) DeepFake, (c) SSIM difference mask showing fake pixels, (d) GradCAM output of a baseline model, (e) GradCAM output of Face-Cutout trained model.}
\label{fig.cam}
\end{figure}

\section{Conclusion}
\label{sec:conclusion}


In this paper, we identified data oversampling as a primary reason why deepfake detection models are not able to generalize well to external data.
We showed the use of face clustering to identify the shortcomings of deepfake datasets and proposed Face-Cutout, a data augmentation method for training convolutional neural networks to overcome these problems. 
Our analysis provides significant directions to evaluate a DeepFake dataset, including a general pre-processing guideline to mitigate overfitting and data leakage. 
Our proposed augmentation policy improves the variation of training data enabling the networks in improving generalizability and robustness to perturbations. We have performed extensive verification to prove that our method is independent of any single type of dataset and performs equally well for multiple types of architectures. We showed that our method improves the deepfake detection performance of existing architectures by $15.2\%$ to $35.3\%$, demonstrating our proposed method's generalizability. Furthermore, our data augmentation technique can be introduced into any existing DeepFake detection pipeline without any significant modifications. 
In the future, we wish to explore the use of this augmentation policy on more diverse face manipulation and forgery datasets.

{\small
\bibliographystyle{ieee_fullname}
\bibliography{main}
}

\newpage
\section*{\centering \Large Appendix}
\medskip
\subsection*{A1. Implementation Details}

\noindent The basic algorithm for Face-Cutout is similar for both sensory group and convex-hull cutout. This is shown in Algorithm \ref{algo}. In both cases, after selecting the specified point group, random polygons are generated by shuffling the points, and each time $\rho$ is calculated to select the polygon with the minimum overlap. The main distinguishing factor is the selection of the polygon region for either sensory group or convex-hull cutout.
\SetKwInOut{Input}{Input}
\SetKwInOut{Output}{Output}

\begin{algorithm}
\caption{Face-Cutout}
    \Input{
        Fake Image $I$\\
        Original Image $O$ (Optional)\\
        Erasing probability $p$\\
        Landmark Coordinates $L$\\
        Cutout threshold $\Gamma_h$\\
        Cutout Fill $F$
    }
    \Output{Augmented Image $I^*$ }
    \label{algo}
    \If{$O \textup{ is \textbf{not empty}}$}{
        Difference Mask $M \gets \textup{diff$(I,O)$}$
    }
    $p_1 \leftarrow Rand(0,1)$\;
    $I^* \leftarrow I$\;
    \uIf{$p_1\geq p$}{
        \textbf{return} $I^*$\;
    }
    \Else{
        $choice \leftarrow Rand[0,1,2,3]$\\
        \lIf{choice $= 0$}{
            $points \leftarrow L[36-47]$
        }
        \lElseIf{choice $= 1$}{
            $points \leftarrow L[48-67]$
        }
        \lElseIf{choice $= 2$}{
            $points \leftarrow L[27-35]$
        }
        \lElse{
            $points \leftarrow L[0-26]$
        }
        \For{$i\gets0$ \KwTo $5$}{
            $I_c \leftarrow \textup{pixels inside the polygon created by}$ \\ 
            \hspace{2.4em}{randomly shuffling the vertices in \\ \hspace{2.2em} \textit{points}}\;
            \If{$O \textup{ is \textbf{empty}}$}{
                $I(I_c) \gets F$\;
                $I^* \leftarrow I$\;
                \textbf{return} $I^*$\;
            }
            $|C_0| \gets \textup{count of $1$'s in } I_c$\;
            $|A| \gets \textup{count of $1$'s in } M$\;
            $\rho \gets \frac{\mid C_o \mid}{\mid A \mid}$\;
            \If{$\rho \leq \Gamma_h$}{
                $I(I_c) \gets F$\;
                $I^* \leftarrow I$\;
                \textbf{return} $I^*$\;
            }
        }
        \textbf{return} $I^*$\;
    }
\end{algorithm}

The calculation of $\rho$ is dependent on the original image since that is used to determine the difference mask. For real faces, the original image is not present. If the original image is not supplied, then this calculation is omitted. 

After selecting the points for the specified group for sensory group cutout, we draw a line using the terminal vertices. The line is drawn between the points: 1) $36, 45$ for eyes, 2) $27, 33$ for nose, and 3) $48, 54$ for the mouth when using DLib. Our implementation used the MTCNN landmark detector that also generates the same terminal points for each landmark. We draw the line using \texttt{cv2.line} function of \texttt{python-opencv} package. Next we used \texttt{scipy.ndimage.binary-dilation} from the \texttt{scipy} package for $5$ iterations with various weights to expand the line and generate a polygon. Each time $\rho$ is measured for the resulting polygon, and we select the polygon with minimum $\rho$.

\vspace{0.2em}
For convex-hull cutout, we use three polygon selections. 
\begin{enumerate}[leftmargin=*]
    \item Firstly, $8$ to $15$ points are selected randomly from the $27$ boundary points. A polygon is drawn using the selected points using the \texttt{skimage.draw.polygon} function from \texttt{skimage} package. For each polygon, we calculate it's area using the equation \ref{eq},
    
    \begin{equation}
    \mathbf{A}=\frac{1}{2} \Bigg| \sum_{i=1}^{n-1} x_{i} y_{i+1}+x_{n} y_{1}-\sum_{i=1}^{n-1} x_{i+1} y_{i}-x_{1} y_{n} \Bigg|
    \label{eq}
    \end{equation}
    
    where, $(x,y)$ are the coordinates of each vertex point. We select the polygon with the maximum area having $\rho \leq \Gamma_h$.
    
    \item We select a number $i$, randomly between $8$ and $15$. We select $i$ consecutive points starting from $0$ till $26$. For example, for $i=11$ on first iteration we select points $0-10$, on second iteration $1-11$, on third iteration $2-12$ and so on. Each time we draw the polygon using the selected points and calculate $\rho$, and it's area. For all polygons that satisfy $\rho \leq \Gamma_h$, we select the one with the maximum area.
    
    \item We draw the polygon using all $27$ boundary points. We find the centroid of this polygon using \texttt{skimage.measure.centroid} function. Then we split the entire polygon laterally and vertically through this centroid, which gives us four sub polygons. From each of these sub-polygons, we select the one with minimum $\rho$.
    
\end{enumerate}

\subsection*{A2. Data Split}
For training and testing on each of the datasets, we used the number of videos given in Table \ref{tab.split}. Since DFDC is magnitudes larger than FF++ and Celeb-DF, for selecting the combined dataset, we used a smaller number of videos from DFDC.

\begin{table}[!ht]
\centering
\resizebox{0.95\columnwidth}{!}{%
\begin{tabular}{c|c|c|c|c|c} 
\hline
\multicolumn{2}{c|}{\multirow{2}{*}{Dataset}} & \multicolumn{2}{c|}{\begin{tabular}[c]{@{}c@{}}Individual\\data split \end{tabular}} & \multicolumn{2}{c}{\begin{tabular}[c]{@{}c@{}}Combined\\ data split \end{tabular}}  \\ 
\cline{3-6}
\multicolumn{2}{c|}{}                         & Real   & Fake                                                                        & Real  & Fake                                                                                                      \\ 
\hline
\hline
\multirow{2}{*}{FF++}     & Train             & 789    & 2,411                                                                       & 600   & 2,000                                                                                                     \\ 
\cline{2-2}
                          & Test              & 100    & 300                                                                         & 200   & 173                                                                                                       \\ 
\hline
\multirow{2}{*}{Celeb-DF} & Train             & 449    & 4,436                                                                       & 300   & 2,000                                                                                                     \\ 
\cline{2-2}
                          & Test              & 60     & 367                                                                         & 209   & 237                                                                                                       \\ 
\hline
\multirow{2}{*}{DFDC}     & Train             & 16,148 & 22,510                                                                      & 5,000 & 7,000                                                                                                    \\ 
\cline{2-2}
                          & Test              & 2,180  & 10,371                                                                      & 2,000 & 4,000                                                                                                     \\
\hline
\end{tabular}
}
\caption{Number of videos in the train and test set for individual testing and combined data testing.}
\label{tab.split}
\end{table}

\end{document}